\documentclass[11pt, a4paper]{article}
\usepackage[hyperref]{acl2017}

\usepackage{times}

\aclfinalcopy 
\def\aclpaperid{3105} 

\usepackage{amsmath}
\usepackage{graphicx}
\usepackage{pslatex}
\usepackage{latexsym}
\usepackage{xspace}
\usepackage{multirow}
\usepackage{amsfonts}
\usepackage{epstopdf}
\usepackage{caption}
\usepackage{tabularx}
\usepackage{pbox}
\usepackage{color}
\usepackage{xcolor}
\usepackage{url}

\usepackage{xargs} 
\usepackage[colorinlistoftodos,prependcaption,textsize=tiny]{todonotes}
\usepackage{marginnote}

\newcommandx{\todoig}[2][1=]{\todo[inline]{SR: #2}\xspace}
\newcommandx{\todois}[2][1=]{\todo[inline]{SG: #2}\xspace}
\newcommandx{\todoik}[2][1=]{\todo[inline]{FK: #2}\xspace}
\newcommandx{\todoil}[2][1=]{\todo[inline]{ML: #2}\xspace}
\newcommandx{\todog}[2][1=]{\todo[linecolor=red,backgroundcolor=red!25,bordercolor=red,#1]{SG: #2}\xspace}
\newcommandx{\todos}[2][1=]{\todo[linecolor=green,backgroundcolor=green!25,bordercolor=green,#1]{GR: #2}\xspace}
\newcommandx{\todok}[2][1=]{\todo[linecolor=cyan,backgroundcolor=cyan!25,bordercolor=cyan,#1]{FK: #2}\xspace}
\newcommandx{\todol}[2][1=]{\todo[linecolor=blue,backgroundcolor=blue!10,bordercolor=blue,#1]{ML: #2}\xspace}

\newcommand \ignore[1]{}

\newcommand{\figref}[2][]{Figure#1~\ref{#2}\xspace}
\newcommand{\tabref}[2][]{Table#1~\ref{#2}\xspace}

\newcommand{\dataset}[1]{#1\xspace}

\newcommand{\coco}{\dataset{MSCOCO}}
\newcommand{\tuhoi}{\dataset{TUHOI}}

\newcommand{\ppmi}{\dataset{PPMI}}
\newcommand{\hico}{\dataset{HICO}}

\newcommand{\imsitu}{\dataset{imSitu}}

\newcommand{\Verse}{\dataset{VerSe}}
\newcommand{\vcoco}{\dataset{VCOCO-SRL}}

\title{An Analysis of Action Recognition Datasets for\\ Language and Vision Tasks}

\author{
Spandana Gella \and Frank Keller\\
Institute for Language, Cognition and Computation \\
School of Informatics, University of Edinburgh \\
10 Crichton Street, Edinburgh EH8 9AB \\
S.Gella@sms.ed.ac.uk, keller@inf.ed.ac.uk
}

\begin{document}
\maketitle

\begin{abstract}

A large amount of recent research has focused on 
tasks that combine language and vision, resulting in
a proliferation of datasets and methods. One such task is 
action recognition, 
whose applications 
include image annotation, scene understanding and 
image retrieval. 
In this survey, we categorize the 
existing approaches based on how 
they conceptualize this problem  
and provide a detailed review of existing datasets, 
highlighting their diversity as well as advantages and
disadvantages. We focus on recently developed 
datasets which link visual information with
linguistic resources and provide a fine-grained syntactic and semantic 
analysis of actions in images.

\end{abstract}

\section{Introduction}

\begin{figure}[t]
\setlength{\tabcolsep}{2pt}
\begin{tabular}{cccc}
\includegraphics[height=20mm, width=17.5mm]{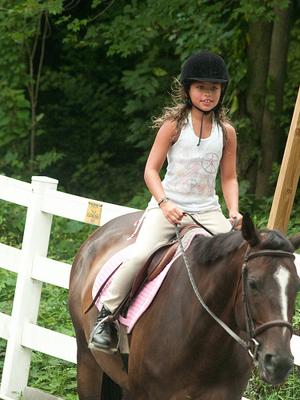} &
\includegraphics[height = 20mm, width=17.5mm]{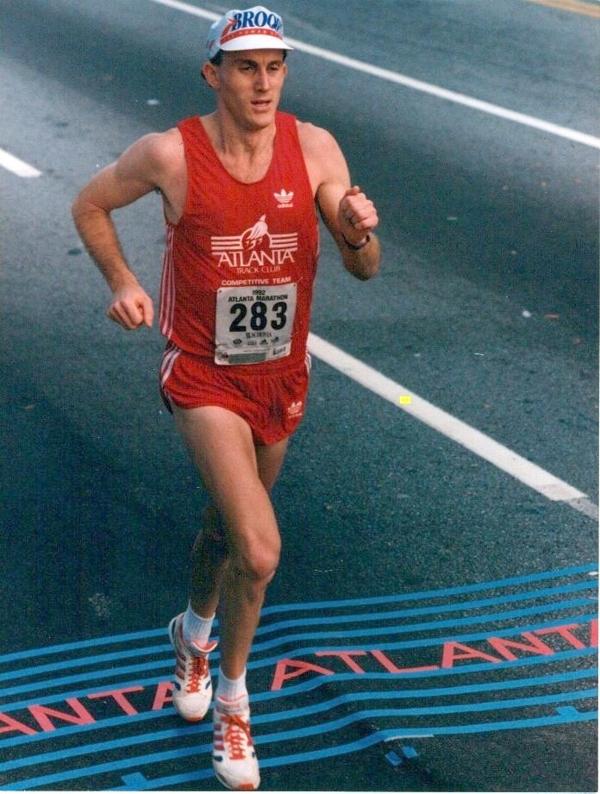} &
\includegraphics[height = 20mm, width=17.5mm]{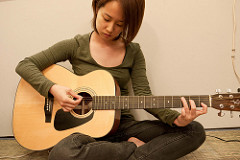}  &
\includegraphics[height = 20mm, width=17.5mm]{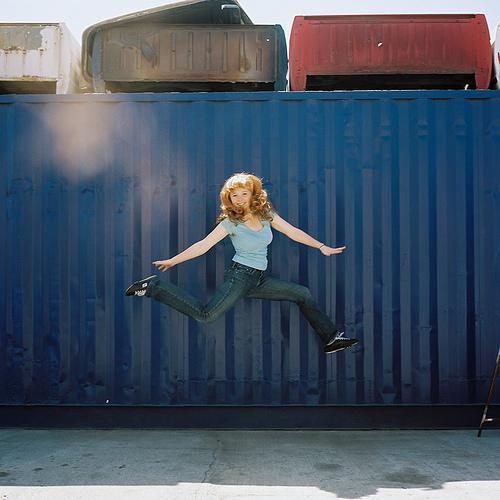} \\
\vspace{-2mm}
\small{riding horse} & 
\small{running} & 
\small{playing guitar} &
\small{jumping}\\
\end{tabular}
\caption{Examples of actions in still images}
\label{fig:example-still-images}
\vspace{-2mm}
\end{figure}

\begin{figure}[t]
\includegraphics[height=50mm]{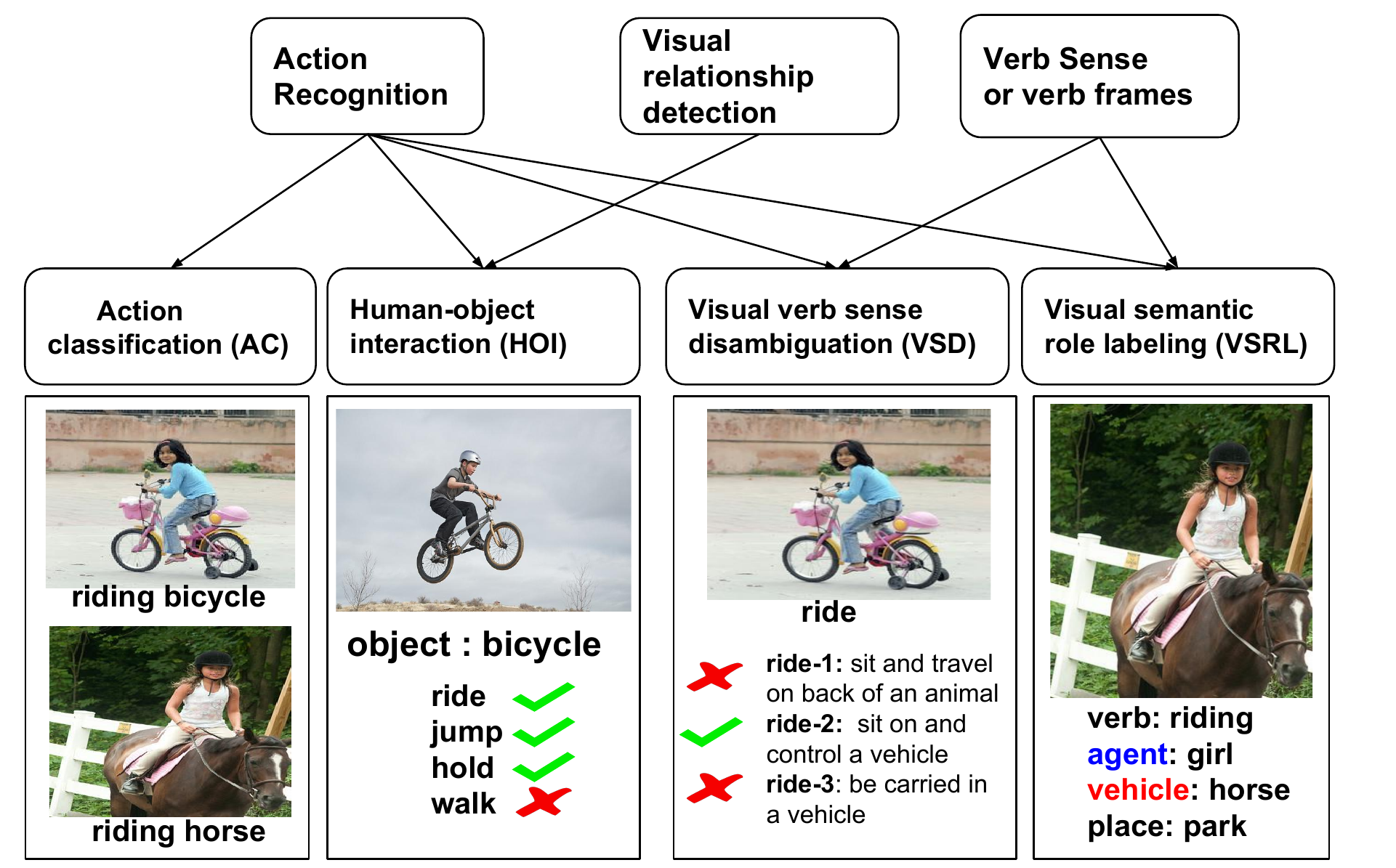}
\caption{Categorization of action recognition tasks in images}
\label{fig:task-categorization}
\end{figure}

Action recognition is the task of identifying 
the action being depicted in a video or still image.
The task is useful for a range of applications such as 
generating descriptions, image/video retrieval, 
surveillance, and human--computer interaction. It has 
been widely studied in 
computer vision, often on videos 
\cite{traffic:actions:94,human:motion:2005}, 
where motion and temporal information provide 
cues for recognizing actions \cite{spatial:temporal:features:lecun:2010}.
However, many actions are recognizable from still 
images, see the examples in 
\figref{fig:example-still-images}.
Due to the absence of motion cues and temporal features \cite{ikizler:2008} action recognition from stills 
is more challenging. 
Most of the existing work can be categorized 
into four tasks: (a)~action classification (AC); 
(b)~determining human--object interaction (HOI); 
(c)~visual verb sense disambiguation (VSD); and
(d)~visual semantic role labeling (VSRL).
In \figref{fig:task-categorization} we illustrate each of these tasks
and show how they are related to each other.

Until recently, action recognition was studied as 
action classification on small-scale datasets 
with a limited number of predefined actions labels
\cite{ikizler:2008,sportsDataset:2009,ppmi:2010,pascal:2010,stanford40:2011}. Often the labels in action 
classification tasks are verb 
phrases or a combination of verb and object 
such as \textit{playing baseball}, \textit{riding horse}.  
These datasets have helped in building models 
and understanding which aspects of an image 
are important for classifying actions, but most methods 
are not scalable to larger numbers of actions 
\cite{google:27kactions:2015}.
Action classification models are trained on 
images annotated with mutually exclusive 
labels, i.e., the assumption is that only a
single label is relevant for a given image. This ignores the fact  
that actions such as \textit{holding bicycle} 
and \textit{riding bicycle} can co-occur in the same image.  
To address these issues and also to understand the range of
possible interactions between humans and objects, the 
human--object interaction (HOI) detection task has been 
proposed, in which all possible interactions between 
a human and a given object have to be identified 
\cite{tuhoi:2014,hico:2015,visual:rel:2016}.

However, both action classification and HOI detection do not consider
the ambiguity that arises when verbs are used as labels, e.g., the verb
\textit{play} has multiple meanings in different contexts. On the
other hand, action labels consisting of verb-object pairs can miss
important generalizations: \textit{riding horse} and \textit{riding
  elephant} both instantiate the same verb semantics, i.e.,
\textit{riding animal}. Thirdly, existing action labels miss
generalizations across verbs, e.g., the fact that \textit{fixing bike}
and \textit{repairing bike} are semantically equivalent, in spite of the
use of different verbs. These observations have led authors to argue
that actions should be analyzed at the level of verb
senses. \citet{Gella2016} propose the new task of visual verb sense
disambiguation (VSD), in which a verb--image pair is annotated with a
verb sense taken from an existing lexical database (OntoNotes in this
case).  While VSD handles distinction between different verb senses,
it does not identify or localize the objects that participate in the
action denoted by the verb. Recent work \cite{vcoco:2015,Yatskar2016}
has filled this gap by proposing the task of visual semantic role
labeling (VSRL), in which images are labeled with verb frames, and the
objects that fill the semantic roles of the frame are identified in
the image.

In this paper, we provide a unified view of action 
recognition tasks, pointing out their strengths 
and weaknesses. We survey existing literature 
and provide insights into existing datasets and models 
for action recognition tasks.

\begin{table*}[t]
\small
\centering
\resizebox{2.0\columnwidth}{!}{
\begin{tabular}{l@{}r@{~}r@{~}r@{~}r@{~}r@{~}r@{~}r@{~}r@{~}c@{~}p{1.1cm}p{4.2cm}} 
\hline
Dataset & Task & \#L &\#V &  Obj & Imgs & Sen & Des & Cln & ML &  Resource & Example Labels \\ 
\hline
Ikizler \cite{ikizler:2008} &  AC &6 & 6 & 0 & 467 &N & N & Y & N & $-$ & running, walking \\ 
Sports Dataset \cite{sportsDataset:2009} & AC & 6 & 6 & 4 & 300 & N & N & Y& N & $-$& tennis serve, cricket bowling \\ 
Willow \cite{delaitre:2010}& AC & 7&6 & 5 &986 & N & N & Y & Y & $-$& riding bike, photographing \\ 
\ppmi \cite{ppmi:2010}  & AC &24 & 2 & 12 & 4.8k & N & N & Y & N &  $-$& play guitar, hold violin  \\ 
Stanford 40 Actions \cite{stanford40:2011}  & AC &40 & 33 & 31 & 9.5k & N & N & Y & N &  $-$& cut vegetables, ride horse \\ 
PASCAL 2012 \cite{pascal:2015} & AC & 11 & 9 & 6 & 4.5k & N & N & Y & Y &  $-$ & riding bike, riding horse \\
89 Actions \cite{89actions:2013} & AC &89 & 36 & 19 & 2k & N & N & Y & N &  $-$ & ride bike, fix bike \\
MPII Human Pose \cite{mpii:pose:2014} &AC & 410 & $-$ & 66 & 40.5k & N & N & Y & N & $-$& riding car, hair styling  \\
\tuhoi \cite{tuhoi:2014}  & HOI & 2974 &$-$ & 189 & 10.8k & N & N & Y & Y &  $-$&sit on chair, play with dog \\
COCO-a \cite{cocoa:2015} & HOI  & $-$ &140 & 80 & 10k & N & Y & Y & Y & VerbNet& walk bike, hold bike \\ 
Google Images \cite{google:27kactions:2015}  & AC  & 2880&$-$ & $-$ & 102k& N & N & N & N &  $-$& riding horse, riding camel \\
\hico \cite{hico:2015} & HOI  & 600 & 111& 80 &  47k & Y & N& Y  &Y &  WordNet &ride\#v\#1 bike; hold\#v\#2 bike \\
VCOCO-SRL \cite{vcoco:2015}& VSRL  & $-$&26 & 48 & 10k & N & Y & Y & Y &  $-$& verb: hit; instr: bat; obj: ball \\ 
\imsitu \cite{Yatskar2016} & VSRL & $-$  &  504& 11k & 126k & Y & N & Y & N & FrameNet  \newline WordNet& verb: ride; agent: girl\#n\#2 \newline vehicle: bike\#n\#1; \newline place: road\#n\#2 \\
\Verse \cite{Gella2016}  &VSD  & 163 & 90& $-$& 3.5k & Y  & Y & Y & N &  OntoNotes& ride.v.01, play.v.02 \\ 
Visual Genome \cite{visual:genome:2016} & VRD & 42.3k & $-$ & 33.8k & 108k & N & N & Y & Y & $-$ & man playing frisbee \\ 
\hline
\end{tabular}
}
\caption{Comparison of various existing action recognition datasets.
 \#L denotes number of action labels in the dataset;
  \#V denotes number of verbs covered in the dataset; 
  Obj indicates number of objects annotated; Sen indicates whether sense
  ambiguity is explicitly handled; Des indicates whether image
  descriptions are included; Cln indicates whether dataset is manually verified; 
  ML indicates the possibility of multiple labels per image; Resource indicates linguistic resource used to label actions.}
\label{tab:action-dataset-stats}
\end{table*}

\section{Datasets for Action Recognition}

We give an overview of commonly used datasets for action recognition
tasks in \tabref{tab:action-dataset-stats} and group them according to
subtask. We observe that the number of verbs covered in these datasets
is often smaller than the number of action labels reported (see \tabref{tab:action-dataset-stats}, columns \#V and \#L) and in many
cases the action label involves object reference. A few of the first
action recognition datasets such as the Ikizler and Willow datasets \cite{ikizler:2008,delaitre:2010}
had action labels such as \textit{throwing} and \textit{running}; they
were taken from the sports domain and exhibited diversity in camera
view point, background and resolution.  Then datasets were created to
capture variation in human poses in the sports domain for actions such
as \textit{tennis serve} and \textit{cricket bowling}; typically
features based on poses and body parts were used to build models
\cite{sportsDataset:2009}.  Further datasets were created based on the
intuition that object information helps in modeling action recognition
\cite{event:classification:li:2007,ikizler:object:scene:2010}, which
resulted in the use of action labels such as \textit{riding horse} or
\textit{riding bike} \cite{pascal:2010,stanford40:2011}.
Not only were most of these datasets domain specific, but the labels
were also manually selected and mutually exclusive, i.e.,
two actions cannot co-occur in the same image. Also, most of
these datasets do not localize objects or identify their semantic
roles.

\subsection{Identifying Visual Verbs and Verb Senses}

The limitations with early datasets (small scale, domain specificity,
and the use of ad-hoc labels that combine verb and object) have been
recently addressed in a number of broad-coverage datasets that offer
linguistically motivated labels. Often these datasets use existing
linguistic resources such as VerbNet \cite{verbnet:2005}, OntoNotes
\cite{ontonotes:2006} and FrameNet \cite{framenet:1998} to classify
verbs and their senses.  This allows for a more general, semantically
motivated treatment of verbs and verb phrases, and also takes into
account that not all verbs are depictable.  For example, abstract
verbs such as \textit{presuming} and \textit{acquiring} are not
depictable at all, while other verbs have both depictable and
non-depictable senses: \textit{play} is non-depictable in
\textit{playing with emotions}, but depictable in \textit{playing
  instrument} and \textit{playing sport}.  The process of identifying
depictable verbs or verb senses is used by \citet{cocoa:2015},
\citet{Gella2016} and \citet{Yatskar2016} to identify visual verbs,
visual verb senses, and the semantic roles of the participating
objects respectively.
In all the cases the process of identifying visual verbs or senses is
carried out by human annotators via crowd-sourcing platforms.  
Visualness labels for 935 OntoNotes verb senses corresponding to 154
verbs is provided by \citet{Gella2016}, while \citet{Yatskar2016}
provides visualness labels for 9683 FrameNet verbs.

\subsection{Datasets Beyond Action Classification}

Over the last few years tasks that combine language and vision such as
image description and visual question answering have gained much
attention. This has led to the creation of new, large datasets such as
\coco \cite{coco:caption:challenge:2015} and the VQA dataset
\cite{vqa:2015}. Although these datasets are not created for
action recognition, a number of attempts have been made to use the
verbs present in image descriptions to annotate actions. The COCO-a,
\Verse and VCOCO-SRL datasets all use the \coco image descriptions to
annotate fine-grained aspects of interaction and semantic roles.

\paragraph{HICO:} The \hico dataset has 47.8k
images annotated with 600 categories of human-object 
interactions with 111 verbs applying to 
80 object categories of \coco. 
It is annotated to include diverse 
interactions for objects and has an average 
of 6.5 distinct interactions per object category. 
Unlike other HOI datasets such as \tuhoi which label interactions as 
verbs and ignore senses, the HOI categories of \hico 
are based on WordNet \cite{WordNet:Miller:1995} verb 
senses. The \hico dataset also has multiple annotations 
per object and it incorporates the information that certain 
interactions such as \textit{riding a bike} and \textit{holding a bike} 
often co-occur. However, it fails to include 
annotations to distinguish between multiple senses of a 
verb.

\paragraph{Visual Genome:} The dataset created by 
\citet{visual:genome:2016} has dense annotations 
of objects, attributes, and relationships between objects. 
The Visual Genome dataset contains 105k images with 40k unique 
relationships between objects. 
Unlike other HOI datasets such as 
\hico, visual genome relationships also include 
prepositions, comparative 
and prepositional phrases such as \textit{near} and 
\textit{taller than}, making the visual relationship task more generic than action recognition. \citet{visual:genome:2016} 
combine all the annotations of objects, relationships, and 
attributes into directed graphs known as scene graphs.

\paragraph{COCO-a:} \citet{cocoa:2015} present Visual VerbNet (VVN), 
a list of 140 common visual verbs manually mined from 
English VerbNet \cite{verbnet:2005}.
The coverage of visual verbs in this dataset is not complete, as many
visual verbs such as \textit{dive}, \textit{perform} and
\textit{shoot} are not included. This also highlights a bias in this
dataset as the authors relied on occurrence in \coco as a verification
step to consider a verb as visual. They annotated 10k images
containing human subjects with one of the 140 visual verbs, for 80 \coco objects. 
This dataset has better coverage of human-object interactions than the \hico dataset despite of missing many visual
verbs.

\paragraph{\Verse:} \citet{Gella2016} created a 
dataset of 3.5k images sampled from the \coco and \tuhoi 
datasets and annotated it with 90 verbs and their
OntoNotes senses to distinguish different verb senses 
using visual context.
This is the first dataset that aims to annotate all 
visual senses of a verb. However, 
the total number of images annotated and number of images for 
some senses is relatively small, which makes it difficult to 
use this dataset to train models. 
The authors further divided their 90 verbs into motion and non-motion 
verbs according to \citet{levin:1993} verb classes and analyzed 
visual ambiguity in the task of visual sense disambiguation.

\paragraph{VCOCO-SRL:} \citet{vcoco:2015} annotated a 
dataset of 16k person instances in 10k images
with 26 verbs and associated objects in the
scene with the semantic roles for each action. The main aim of 
the dataset is to build models for visual 
semantic role labeling in images. This task involves identifying the
actions depicted in an image, along with the people and objects that
instantiate the semantic roles of the actions. 
In the \vcoco dataset, each person instance is annotated with 
a mean of 2.8 actions simultaneously. 

\paragraph{\imsitu:} \citet{Yatskar2016} annotated a large dataset of 
125k images with 504 verbs, 1.7k semantic roles and 
11k objects. They used FrameNet verbs, 
frames and associated objects or scenes with roles to develop the dataset. 
They annotate every image with a single verb and the semantic roles of
the objects present in the image. VCOCO-SRL the is dataset most similar
to \imsitu, however VCOCO-SRL includes localization information of
agents and all objects and provides multiple action annotations per
image. On the other hand, \imsitu is the dataset that covers
highest number of verbs, while also omitting many commonly studied
poly\-semous verbs such as \textit{play}.

\subsection{Diversity in Datasets}

With the exception of a few datasets such as COCO-a, \Verse, \imsitu
all action recognition datasets have manually picked labels or focus
on covering actions in specific domains such as sports. Alternatively,
many datasets only cover actions relevant to specific object
categories such as musical instruments, animals and vehicles. In the
real world, people interact with many more objects and perform actions
relevant to a wide range of domains such as personal care, household
activities, or socializing. This limits the diversity and coverage of
existing action recognition datasets.  Recently proposed datasets
partly handle this issue by using generic linguistic resources to
extend the vocabulary of verbs in action labels. The diversity 
issue has also been highlighted and addressed in recent video 
action recognition datasets,
\cite{activitynet:2015,charades:2016}, 
which include generic household activities.  An analysis of 
various image description and question answering datasets 
by \citet{vl-caption-survey:2015} shows
the bias in the distribution of word categories. Image description
datasets have a higher distribution of nouns compared to other word
categories, indicating that the descriptions are object specific,
limiting their usefulness for action-based tasks.

\section{Relevant Language and Vision Tasks}

Template based description generation systems for both videos and
images rely on identifying subject--verb--object triples and use
language modeling to generate or rank descriptions
\cite{corpus:description:generation:yang:2011,videos:captions:2014,Bernardi:ea:16}. Understanding actions also plays an important 
role in question answering, especially when the question 
is pertaining to an action depicted in the image. 
There are some specifically curated question
answering datasets which target human activities or relationships
between a pair of objects \cite{madlib:2015}. \citet{mallya2016learning} have shown that systems trained on action recognition datasets 
could be used to improve the accuracy of visual question answering systems that handle questions related to human activity and human--object relationships.
Action recognition datasets could be used to learn actions that are visually similar such as
\textit{interacting with panda} and \textit{feeding a panda} or
\textit{tickling a baby} and \textit{calming a baby}, which cannot be
learned from text alone \cite{google:27kactions:2015}. Visual semantic role labeling is a crucial step for 
grounding actions in the physical world \cite{Yang:Grounded:SRL:2016}.

\section{Action Recognition Models}

Most of the models proposed for action classification and
human--object interaction tasks rely on identifying higher-level
visual cues present in the image, including human bodies or body parts
\cite{ikizler:2008,sportsDataset:2009,stanford40:2011,mpii:pose:2014},
objects \cite{sportsDataset:2009}, and scenes
\cite{event:classification:li:2007}. Higher-level visual cues are
obtained through low-level features extracted from the image such as
Scale Invariant Feature Transforms (SIFT), Histogram of Oriented
Gradients (HOG), and Spatial Envelopes (Gist) features
\cite{lowe1999object,dalal2005histograms}. These are useful in
identifying key points, detecting humans, and scene or background
information in images, respectively.  In addition to identifying
humans and objects, the relative position or angle between a human and
an object is useful in learning human--object interactions
\cite{tuhoi:2014}. Most of the existing approaches rely on learning
supervised classifiers over low-level features to predict action
labels.

More recent approaches are based on end-to-end convolutional neural
network architectures which learn visual cues such as objects and
image features for action recognition
\cite{hico:2015,zhou2016learning,mallya2016learning}. While most of
the action classification models rely solely on visual information,
models proposed for human--object interaction or visual relationship
detection sometimes combine human and object identification (using
visual features) with linguistic knowledge
\cite{tuhoi:2014,visual:genome:2016,visual:rel:2016}. Other work on
identifying actions, especially methods that focus on relationships
that are infrequent or unseen, utilize word vectors learned on large
text corpora as an additional source of information
\cite{visual:rel:2016}. Similarly, \citet{Gella2016} show that
embeddings generated from textual data associated with images (object
labels, image descriptions) is useful for visual verb sense
disambiguation, and is complementary to visual information.

\section{Discussion}

Linguistic resources such as WordNet, OntoNotes, and FrameNet play a
key role in textual sense disambiguation and semantic role
labeling. The visual action disambiguation and visual semantic role
labeling tasks are extensions of their textual counterparts, where
context is provided as an image instead of as text. Linguistic
resources therefore have to play a key role if we are to make rapid
progress in these language and vision tasks. However, as we have shown
in this paper, only a few of the existing datasets for action
recognition and related tasks are based on linguistic resources
\cite{hico:2015,Gella2016,Yatskar2016}. This is despite the fact that
the WordNet noun hierarchy (for example) has played an important role
in recent progress in object recognition, by virtue of underlying the
ImageNet database, the de-facto standard for this task \cite{imagenet:2014}. The success of
ImageNet for objects has in turn helped NLP tasks such as bilingual
lexicon induction \cite{vulic2016multi}. In our view,
language and vision datasets that are based on the WordNet, OntoNotes,
or FrameNet verb sense inventories can play a similar role for tasks
such as action recognition or visual semantic role labeling, and
ultimately be useful also for more distantly related tasks such as
language grounding.

Another argument for linking language and vision datasets with
linguistic resources is that this enables us to deploy the datasets in
a multilingual setting. For example a polysemous verb such as
\textit{ride} in English has multiple translations in German and
Spanish, depending on the context and the objects involved. Riding a
horse is translated as \emph{reiten} in German and \emph{cabalgar} in
Spanish, whereas riding a bicycle is translated as \emph{fahren} in
German and \emph{pedalear} in Spanish. In contrast, some polysemous
verb (e.g., English \emph{play}) are always translated as the same
verb, independent of sense (\emph{spielen} in German). Such sense
mappings are discoverable from multilingual lexical resources (e.g.,
BabelNet, \citealt{babelnet:2010}), which makes it possible to
construct language and vision models that are applicable to multiple
languages. This opportunity is lost if language and vision dataset are
constructed in isolation, instead of using existing linguistic
resources.

\section{Conclusions}

In this paper, we have shown the evolution of action recognition
datasets and tasks from simple ad-hoc labels to the fine-grained
annotation of verb semantics.  It is encouraging to see the recent
increase in datasets that deal with sense ambiguity and annotate
semantic roles, while using standard linguistic resources. One major
remaining issue with existing datasets is their limited coverage, and
the skewed distribution of verbs or verb senses. Another challenge is
the inconsistency in annotation schemes and task definitions across
datasets. For example \citet{hico:2015} used WordNet senses as
interaction labels, while \citet{Gella2016} used the more
coarse-grained OntoNotes senses. \citet{Yatskar2016} used FrameNet
frames for semantic role annotation, while \citet{vcoco:2015} used
manually curated roles.  If we are to develop robust, domain
independent models, then we need to standardize annotation schemes and
use the same linguistic resources across datasets.

\newpage
\bibliographystyle{acl_natbib}
\bibliography{references}

\end{document}